\documentclass[aip,cha,reprint]{revtex4-1}
\usepackage{lineno}
\usepackage{graphicx}
\usepackage{natbib}
\usepackage{color}
\usepackage{amssymb}
\usepackage{subfigure}
\usepackage{algorithm2e}
\usepackage{dsfont} 

\begin{document}


\title{Multiscale Fractal Descriptors Applied to Texture Classification}
\author{Jo\~{a}o Batista Florindo}
\affiliation{Universidade de S\~{a}o Paulo \\ Instituto de F\'{i}sica de S\~{a}o Carlos (IFSC) - http://scg.ifsc.usp.br \\ jbflorindo@gmail.com}
\author{Odemir Martinez Bruno}
\affiliation{Universidade de S\~{a}o Paulo \\ Instituto de F\'{i}sica de S\~{a}o Carlos (IFSC) - http://scg.ifsc.usp.br \\ bruno@ifsc.usp.br}

\begin{abstract}
This work proposes the combination of multiscale transform with fractal descriptors employed in the classification of gray-level texture images. We apply the space-scale transform (derivative + Gaussian filter) over the Bouligand-Minkowski fractal descriptors, followed by a threshold over the filter response, aiming at attenuating noise effects caused by the final part of this response. The method is tested in the classification of a well-known data set (Brodatz) and compared with other classical texture descriptor techniques. The results demonstrate the advantage of the proposed approach, achieving a higher success rate with a reduced amount of descriptors.
\end{abstract}


\maketitle


\section{Introduction}

Texture analysis is one of the most studied problems in Computer Vision area \cite{H67,MS98,MBLS01}. The methodology employed in such problems may be divided into 4 categories \cite{MS98}, to know, structural, statistical, spectral and model-based methods. In this last category, we find the fractal techniques.

Fractal geometry shows to be an efficient tool to characterize and discriminate complex objects usually found in texture images \cite{QMACG08,TWZ07,MG05}. This precision of fractal representation is explained by the flexibility of a fractal object, given its nature instrinsically complex. Moreover, fractals are strongly identified by their self-similarity behavior, that is, each part constitutes a copy only linearly transformed of the whole structure. Such characteristic is also widely observed in many objects from the real world, in a greater or smaller degree of intensity.

Among the most known fractal-based techniques, we can cite the multifractals \cite{H01}, the multiscale fractal dimension \cite{MCSM02} and the fractal descriptors. Here, we choose fractal descriptors approach, due to their high efficiency demonstrates in several works studying texture analysis \cite{BPFC08,PPFVOB05}.

Despite its efficiency, fractal descriptors still have problems with the analysis of images presenting a significant amount of patterns and details, which may be observed only at some specific scales. This is explained by the fact that, although fractal descriptors analyzes the image under different scales, the descriptors themselves are represented globally and are not able to highlight any kind of individual scale.

With the aim of overcoming such drawback, this work develops and studies a novel approach where a multiscale transform \cite{W84} is applied to the original descriptors. Particularly, we employed a space-scale transform (derivative + Gaussian filter) and applyed a threshold over the filter response in order to discard some irrelevant information. The application of a multiscale transform enabled the use of fractal descriptors as not only a global curve which describes the image, but a set of informations about details, patterns and irregularities observed at a wide range of scales.

\section{Fractal Geometry}

Fractal geometry is the area which studies fractal objects. Fractals are complex geometrical structures characterized by an infinite degree of self-similarity, that is, each part of the object corresponds to a copy of the whole, only affected by linear geometrical transforms (affine transforms) \cite{M68,PBS88}.

As well as fractals, we may find a large number of structures in nature also characterized by a high level of repetition of patterns. Moreover, we also verify that many of such natural object cannot find a precise enough representation through Euclidean classical geometry \cite{M68,PBS88}. Such observations yield researchers to aplpy fractal geometry tools in a wide range of problems in several areas \cite{QMACG08,TWZ07,MG05}. Among the employed fractal tools, fractal dimension is the most used.

\subsection{Fractal Dimension}

As well as in Euclidean geometry we have metrics like perimeter and area, fractal dimension is a fundamental characteristic measure of a fractal object.

Essentially, this metric is obtained by measuring the extension of the object through a rule with a variable side-length. The dimension of an object $S$ is given by the following expression:
\begin{equation}
\label{eq:dimension}
	D = \lim_{\epsilon \rightarrow 0}\frac{\log(N_{\epsilon}(S))}{\log(\epsilon)},
\end{equation}
where $N$ is the number of units with length $\epsilon$ necessary to cover the whole object.

The above equation is only usable when we know the exact mathematical rule which gave rise to the fractal object. In practical situations like in this work, when we are dealing with objects represented in a digital image, we must turn to estimation methods which provide an approximate fractal dimension value. Most of such methods are based on a generalization of the Equation \ref{eq:dimension}, where, instead of the number of rules $N$, we use any complexity metric whether in spatial or frequency space.

Among such estimation methods, we opted to use the Bouligand-Minkowski (BM) technique. This choice was based on its good performance in texture analysis problems similar to those showed in this work \cite{BCB09,PPFVOB05,BPFC08,FB12,FBCB12}. 

The Bouligand-Minkowski approach consists in some basic steps. Initially, the gray-level intensity texture image $I \in [1:M] \times [1:N] \rightarrow \Re$ is mapped onto a three-dimensional surface $S$ in the following manner:
\begin{equation}
	S = \{i,j,f(i,j)|(i,j) \in [1:M] \times [1:N]\},
\end{equation}
and 
\begin{equation}
	f(i,j) = \{1,2,...,max_{I}\}|f = I(i,j),
\end{equation}
being $max_{I}$ the maximum intensity value of the image.

In the following, the surface is dilated by a sphere with a variable radius $r$, that is, around each point with coordinates $(x,y,z)$ of the surface, we construct a sphere with radius $r$ and center at $(x,y,z)$. For each value of $r$ we can compute the dilation volume $V(r)$ in a strightforward manner by:
\begin{equation}
	V(r) = \{\bigcup_{p \in S}B_r(p)\},
\end{equation}
where $B_r$ are the spheres. An equivalent expression is the following:
\begin{equation}\label{eq:vol}
	V(r) = \{p_1 \in R^3 | \exists p_2 \in S : |p_1 - p_2| \leq r\},
\end{equation}
where $p_1$ is a point outside the surface, while $p_2$ is inside $S$.

Finally, varying the values of $r$ along a specifical interval obtained empirically, the dimension is estimated through the following expression:
\begin{equation}
	D = 3 - \lim_{\epsilon \rightarrow 0}\frac{\log(V_{r}(S))}{\log(r)}.
\end{equation}

\begin{figure}[!htpb] 
					 \centering
					 \mbox{\subfigure[]{\includegraphics[width=0.125\textwidth]{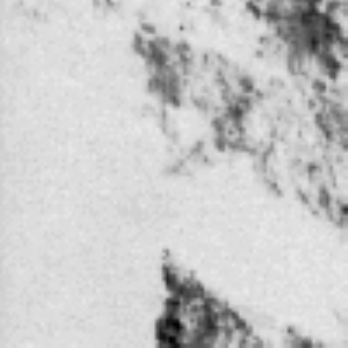}}
					 			 \subfigure[]{\includegraphics[width=0.125\textwidth]{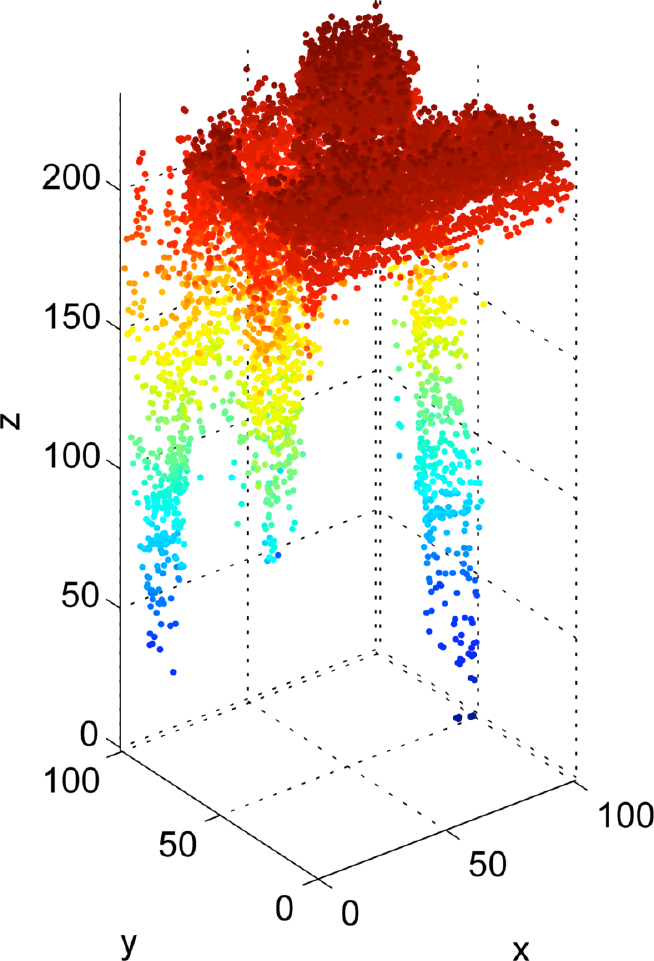}}
					 			 \subfigure[]{\includegraphics[width=0.125\textwidth]{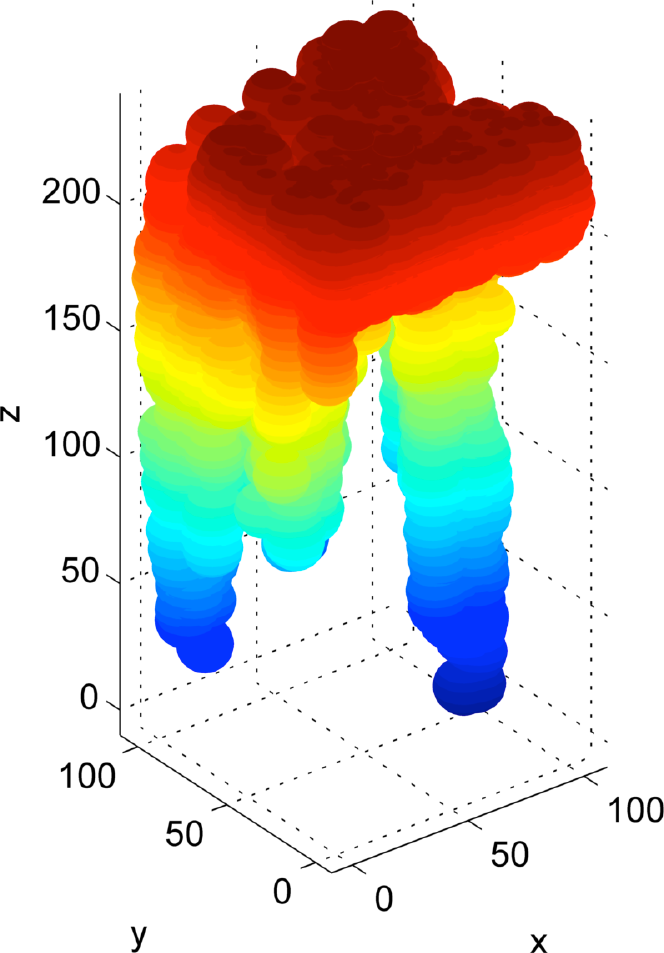}}}
					 \mbox{\subfigure[]{\includegraphics[width=0.25\textwidth]{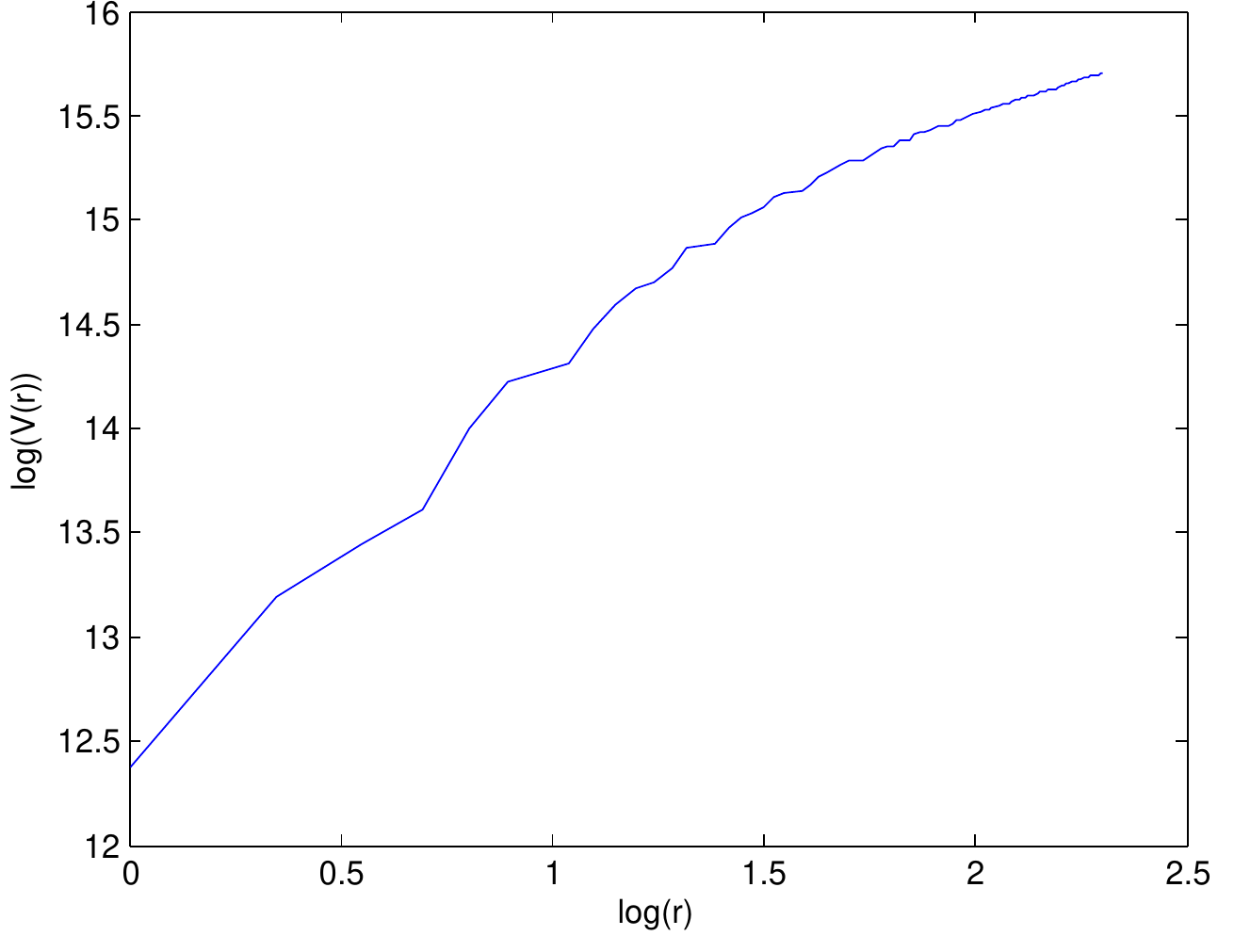}}}					 
           \caption{Dilation of the surface to obtain the Bouligand-Minkowski dimension. (a) Texture image. (b) Surface representation. (c) An example of dilation of the surface (in this case, with $r = 10$). d) Curve $\log(V) \times \log(r)$.}
           \label{fig:dilat}                                  
   \end{figure}

The Figure \ref{fig:dilat} illustrates the dilation process.

\section{Fractal Descriptors}

Despite its importance as a characterizer of objects, fractal dimension is still limitted, mainly when we need to discriminate among a large number of objects with a high degree of similarity. Such difficulty can be justified by two main aspects. The first is that fractal dimension is only a unique value to describe the whole complexity of an object. The second is that we observe a lot of objects with a completely distinct appearance whilst they have the same estimated fractal dimension \cite{FB12}. In order to overcome such drawback, Bruno et al. \cite{BPFC08} developed a novel methodology called fractal descriptors.

This technique consists in composing a set of features (descriptors) from the power-law fractality curve of the analyzed object. In Bouligand-Minkowski method used here, such descriptors are obtained from the dilation volume values. We explain the procedure to compute such values in the following.

Based on the Equation \ref{eq:vol} and on the fact that we are dealing with a discrete space, the Euclidean Distance Transform (EDT) \cite{FCTB08} shows to be a powerful method to compute BM dimension. In this way, we construct a distance mapping for the surface $S$, using pre-defined values for the distance:
\begin{equation}
	E = {0,1,\sqrt{2},...,l,...},
\end{equation}
where
\begin{equation}
	l \in D = \{d | d = (i^2+j^2)^{1/2};i,j \in \mathbb{N}\}.
\end{equation}

Thus, we define the set $g_r(S)$ of points at a distance $r$ from a point with coordinates $(S_x,S_y,S_z)$ in $S$:
\begin{equation}
	 g_r(S) = \left\{(x,y,z)|[(x-S_x)^2 + (y-S_y)^2 \\
	 + (z-S_z)^2]^{1/2} = E(r) \right\}.
\end{equation}

Finally, the dilation volume is given by:
\begin{equation}
	V(r) = \sum_{i=1}^{r}{Q(i)},
\end{equation}
where $Q(i)$ represents the number of points at a distance $i \leq r$:
\begin{equation}
	Q(i) = \sum_{(x,y,z)}{\chi_{g_r}(x,y,z)},
\end{equation}
where $\chi$ is the indicator function: $\chi_A(x) = 1$ if $x \in A$ and $\chi_A(x) = 0$ otherwise.

The descriptors are obtained from the power-law relation $u$ between $V(r)$ and $r$, that is:
\begin{equation}
	u:\log(r) \rightarrow \log(V(r)).
\end{equation}
In this way, $V(r)$ acts as a fractality measure while $r$ corresponds to the representation scale.

The function $u$ can be used directly, as in \cite{BCB09} or after some kind of transform. In this work, we propose a multiscale transform to enhance the results obtained with fractal descriptors.

\section{Proposed Method}

In image analysis, and particularly texture analysis, we are often faced with situations where the behavior of some analysis method varies severely according to the scale at which the image is analyzed. Multiscale transform was developed with the aim of attenuating such difficulty, providing robust information about features of the object which are observed only at specific scales.

The multiscale transform, applied over a curve $u(t)$ like that of fractal descriptors, consists in a mapping function of $u(t)$ onto $U(b,a)$, where $b$ is directly related to $t$ and $a$ is the scale parameter.

Basically, the literature shows three categories of multiscale transform, to know, space-scale, time-frequency and time-scale \cite{CC00}. Here, we adopted the space-scale approach, given its large use in image analysis and particularly interesting results in works related to the identification of patterns in natural objects, like those studied in this work \cite{PPFVOB05,BPFC08}.

In this approach, the function $U(b,a)$ is obtained by derivating the function $u(t)$ and applying a Gaussian smoothing filter $g_a$, where $a$ is the smoothing parameter. More formally, we have:
\begin{equation}
	U(b,a) = U^1(t) * g_a(t) = U^1(t,a),
\end{equation}
where $U^1(t,a)$ is the convolution of the original signal $u(t)$ with the first derivative of the gaussian $g^1_a$, that is:
\begin{equation}
	U^1(t,a) = u(t) * g_a^1(t).
\end{equation}

Inasmuch as multiscale is a two-dimensional transform applied over a one-dimensional signal, it adds a subtantial amount of redundance information. To solve this drawback, we employed a technique based on the adjustment of visual system to select objects in a scene \cite{CC00}. Such solution, named fine-tuning selection consists in selecting a specific value of parameter $a$ and projecting the transform over that point. Here, we select the value of $a$ empirically.

Finally, we applied a threshold over the obtained signal in order to eliminate noises added by the application of the multiscale over the final part of Bouligand-Minkowski curve, where the data becomes more sensible due to the high level of wavefront interference. The threshold point is also determined empirically.

\begin{figure}[!htpb]
\centering
\includegraphics[width=0.5\textwidth]{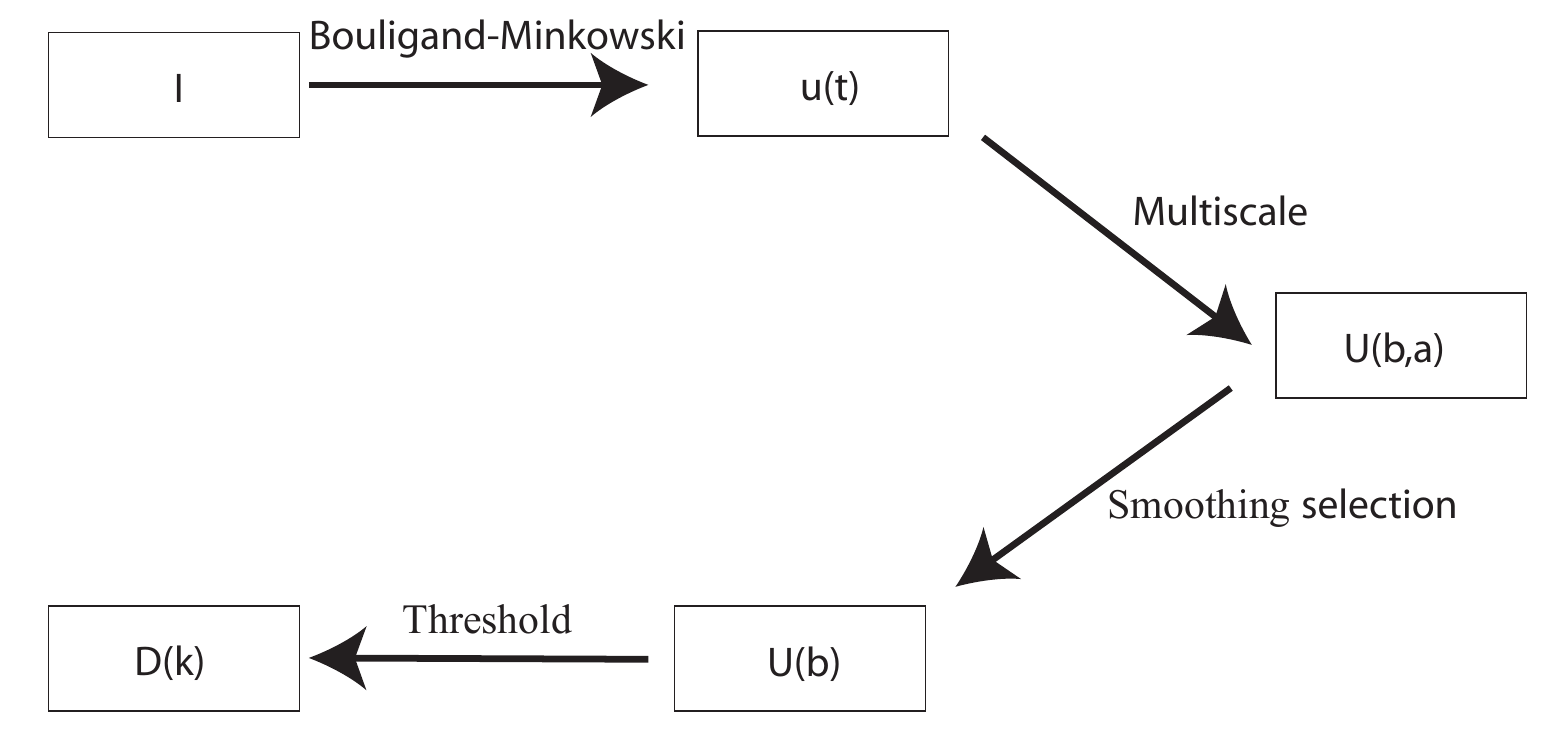}
\caption{A diagram illustrating the essential steps involved in the proposed method.}
\label{fig:method}
\end{figure}

The Figure \ref{fig:method} shows a summary of the steps in the proposed method in a box diagram.

\begin{figure}[!htpb]
\centering
\includegraphics[width=0.5\textwidth]{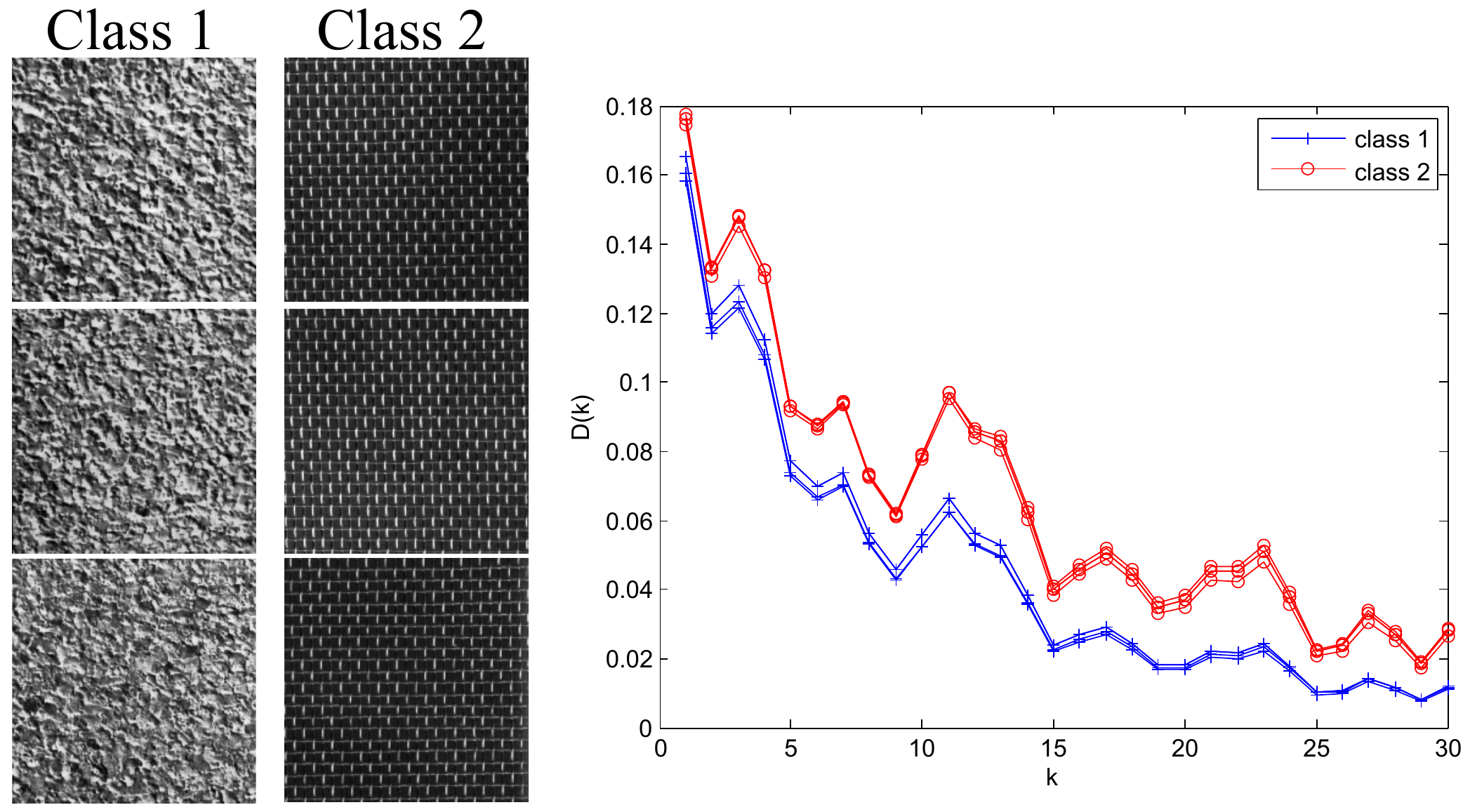}
\caption{Discrimination ability of proposed descriptors $D(k)$.}
\label{fig:discrim}
\end{figure}

The Figure \ref{fig:discrim} depicts the discrimination power of the proposed descriptors, distinguishing visually among samples from two classes.

\section{Experiments}

The efficiency of the proposed method is verified by a comparison with other classical and state-of-the-art methods in the classification of a texture data set. 

Here, we used the well-known Brodatz \cite{B66} texture data. This corresponds to 1110 images, each one divided into 10 windows. Each image is a class and each window is a sample.

The classification itself is carried out by the Linear Discriminant Analysis (LDA) classifier \cite{DH00} using a Hold-Out scheme, with one half of the data to train and the other half to validate.

We compare the success rate to the use of Bouligand-Minkowski descriptors (without multiscale) \cite{BCB09}, Gabor wavelets descriptors \cite{MM96}, Co-occurrence matrix \cite{H67} and a multiscale approach described in \cite{PV02}.

\section{Results}  

The Table \ref{tab:brodatz} shows the success rate of the compared descriptors in the classification task of Brodatz data. We also show the number of descriptors used to provide the best result and the $\kappa$ index \cite{DH00}, a measure of the reliability of the result. The value of $1-\kappa$ may also be used as an error measure of the experiment. For the propoed method, we employed a parameter $a = 0.7$ and a threshold after the 51$^{th}$ element.

We observe that the proposed technique provided the best results with a reasonable number of descriptors. Such aspect is interesting once beyond the computational efficiency, the obtainment of a high success rate using less descriptors turns possible the attenuation of dimensionality curse effects and produce a more reliable result. Also it is important to emphasize the high $\kappa$ value obtained by the multiscale approach, implying in a robust classification process.
\begin{table*}[!htpb]
	\centering
		\begin{tabular}{cccc}
			\hline
                 Method                        & Correctness Rate (\%) & $\kappa$ & Number of descriptors\\
                 \hline
                 \hline
Co-occurrence & 92.0721 & 0.9201 & 84\\
Gabor & 90.0901 & 0.9001 & 20\\
Multifractal & 37.4775 & 0.3732 & 101\\
Minkowski & 98.9189 &  0.9891 & 85\\
Proposed & 99.4595 & 0.9945 & 51\\			
								 \hline			
		\end{tabular}
	\caption{Correctness rate for the Brodatz dataset.}
	\label{tab:brodatz}
\end{table*}
%
%

\begin{figure}[!h] 
					 \centering
					 \mbox{\subfigure[]{\includegraphics[width=0.25\textwidth]{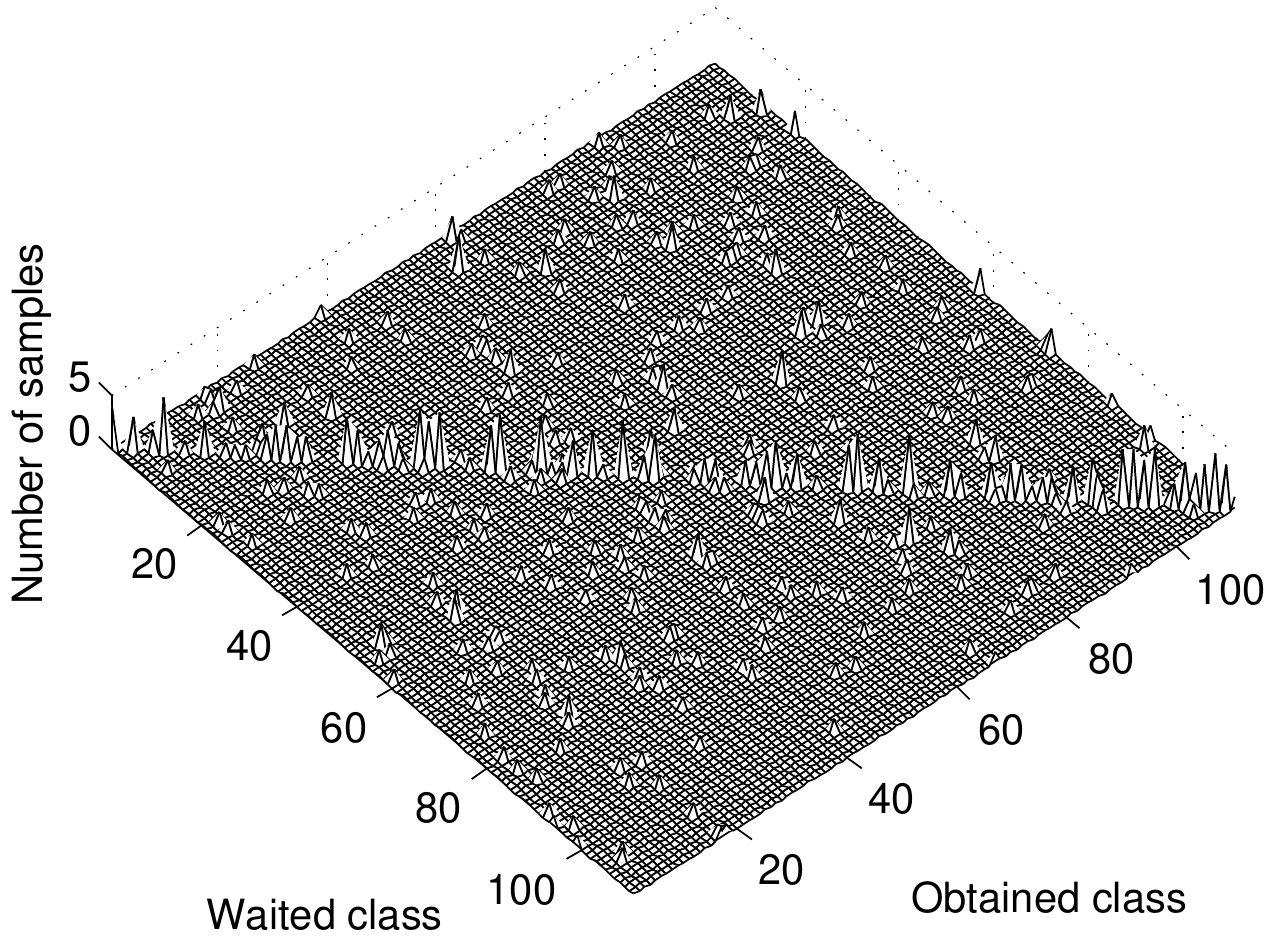}}
					 			 \subfigure[]{\includegraphics[width=0.25\textwidth]{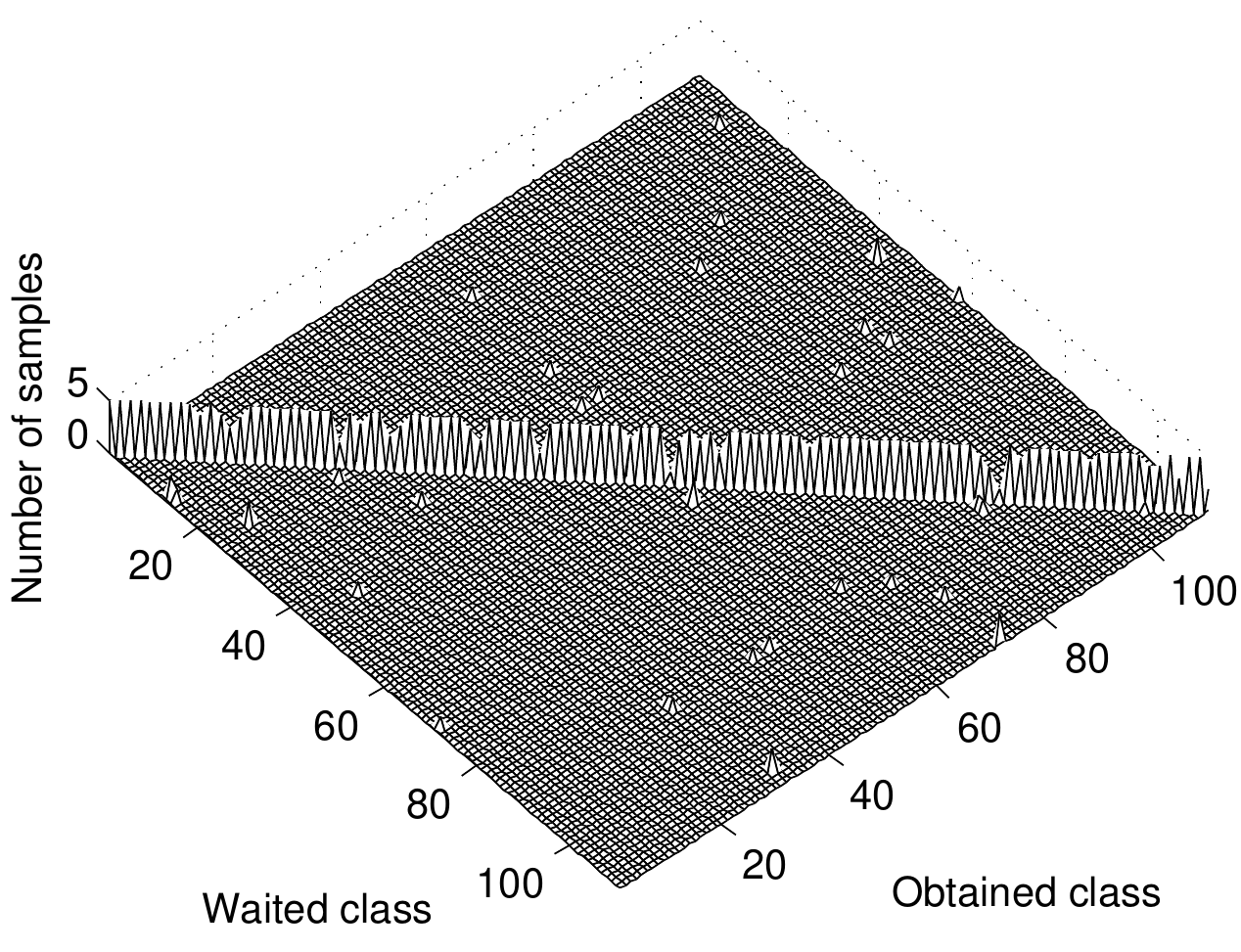}}}
					 \mbox{\subfigure[]{\includegraphics[width=0.25\textwidth]{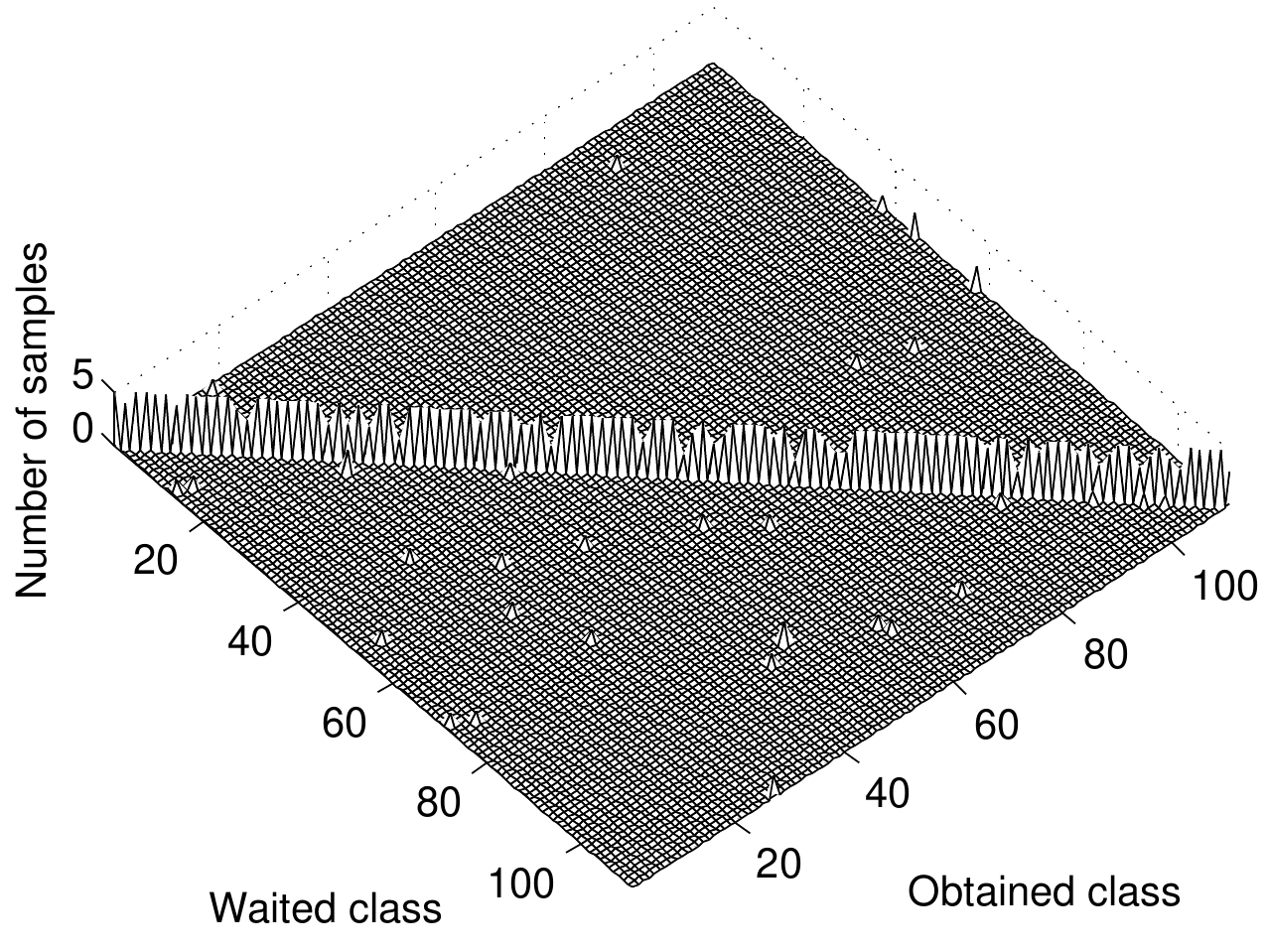}}
								 \subfigure[]{\includegraphics[width=0.25\textwidth]{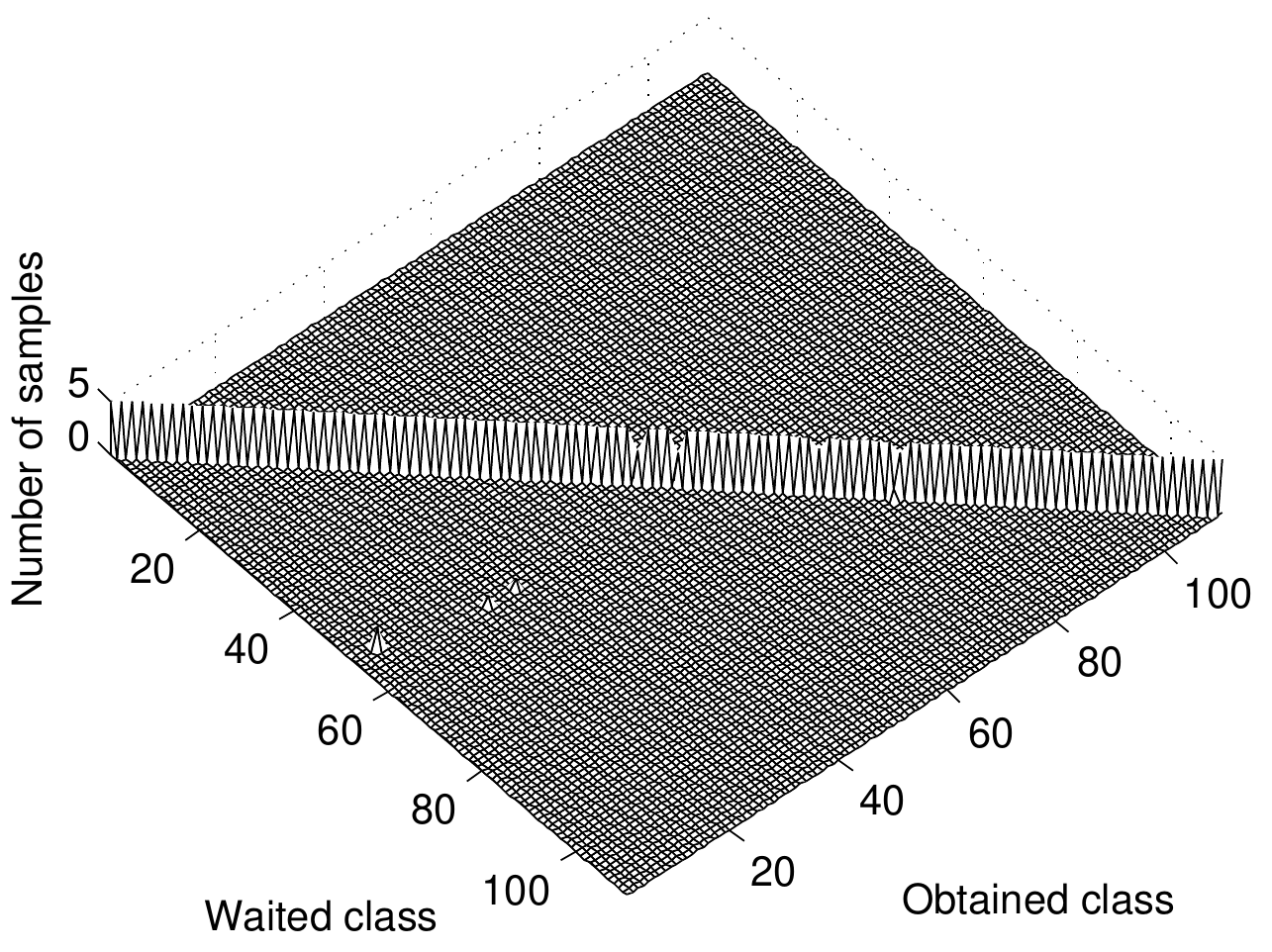}}}	
					 \mbox{\subfigure[]{\includegraphics[width=0.25\textwidth]{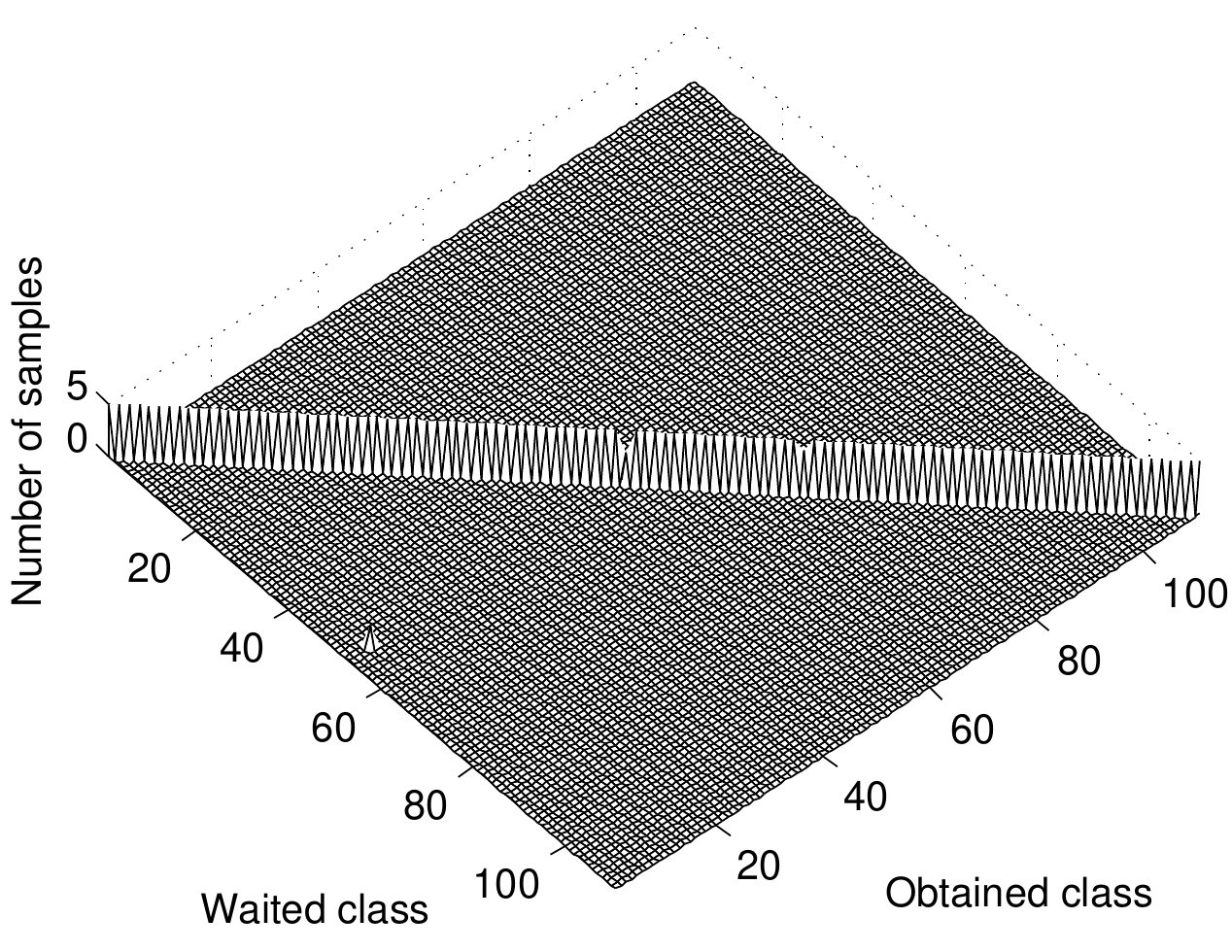}}}								 				 
           \caption{Confusion matrices. (a) Multifractal. (b) Co-occurrence. (c) Gabor. d) Minkowski. e) Proposed method.}
           \label{fig:CM}                                  
   \end{figure}

Complementing the results in the previous table, we show the confusion matrices for each method in the Figure \ref{fig:CM}. In this figure, we have a surface where the height at each point corresponds to the number of samples classified as being from a predicted class (obtained) while pertains to an actual class (waited). In this graphical, we must observe the diagonal (corresponding to the samples classified correctly) and peaks outside the diagonal, corresponding to the misclassifications. We see that such matrices reinforce the Table \ref{tab:brodatz}. The best results correspond to matrices with less ``holes'' inside the diagonal and less peaks outside. For instance, the Mikowski and the proposed method have a similar matrix, except for the fact that Minkowski shows 3 little peaks outside the diagonal, while the proposed technique presents only one.
   
In summary, we observe that the combination of fractal descriptors and space-scale transform constituted a powerful and robust method to provide texture descriptors. Although fractal descriptors provide a good result, as attested by the success rate of Minkowski method, it is still limitted to an analysis more global of the image, which limits its efficiency in more complex situations as, for instance, when we have a high degree of similarity from different classes and/or a high dissimilarity intra-class. The global fractal descriptors approach also has some dificulties in dealing with noised images. The multiscale approach allows the highlighting of scale-dependent nuances, characterizing the image in a richer way, unveiling important structures which only appear at some specific scale as well as isolating artifacts and noises inherent to the original image.

\section{Conclusion}

This work proposed the application of a multiscale transform to the fractal descriptors method for the classification of texture images. We applied the space-scale transform (derivative followed by Gaussian filtering) to the Bouligand-Minkowski fractal descriptors. After that, we also applied a threshold to the filter response to eliminate the less significant information.

We compared the method to other well-known texture descriptors and the results showed that the multiscale combination porvided the best success rate even with a reduced number of descriptors. Such performance confirmed the efficiency of multiscale transform associated to fractal geometry, once it allows the study of details which only can be detected at specific scales, turning, in this way, the analysis more complete and robust.

The results also suggest the application of the proposed approach to other problems from the real world, involving the discrimination of objects in complex scenarios.

\section*{Acknowledgments}

Joao B. Florindo gratefully acknowledges the financial support of CNPq (National Council for Scientific and Technological Development, Brazil) (Grant \#140624/2009-0). Odemir M. Bruno gratefully acknowledges the financial support of CNPq (National Council for Scientific and Technological Development, Brazil) (Grant \#308449/2010-0 and \#473893/2010-0) and FAPESP (The State of S\~ao Paulo Research Foundation) (Grant \# 2011/01523-1).


%

\end{document}